\documentclass{article} % For LaTeX2e
\usepackage[table]{xcolor}
\usepackage{colm2024_conference}
\usepackage{microtype}
\usepackage{hyperref}
\usepackage{url}
\usepackage{booktabs}
\usepackage{amsmath}
\usepackage{multirow}
\usepackage{multicol}
\usepackage{CJKutf8}
\usepackage{tcolorbox}

\definecolor{darkblue}{rgb}{0, 0, 0.5}
\hypersetup{colorlinks=true, citecolor=darkblue, linkcolor=darkblue, urlcolor=darkblue}
\newcommand{\ignore}[1]{}

\title{DianJin-R1: Evaluating and Enhancing Financial Reasoning in Large Language Models}

\author{Jie Zhu$^1$, Qian Chen$^1$, Huaixia Dou$^{1,2}$, Junhui Li$^2$,  \\ \bf{Lifan Guo$^1$, Feng Chen$^1$, Chi Zhang$^1$}\\
$^1$Qwen DianJin Team, Alibaba Cloud Computing\\
$^2$Soochow University\\
}

\colmfinalcopy % Uncomment for camera-ready version, but NOT for submission.
\begin{document}

\maketitle

\begin{CJK*}{UTF8}{gkai}

\begin{abstract}
Effective reasoning remains a core challenge for large language models (LLMs) in the financial domain, where tasks often require domain-specific knowledge, precise numerical calculations, and strict adherence to compliance rules. We propose DianJin-R1, a reasoning-enhanced framework designed to address these challenges through reasoning-augmented supervision and reinforcement learning. Central to our approach is DianJin-R1-Data, a high-quality dataset constructed from CFLUE, FinQA, and a proprietary compliance corpus (Chinese Compliance Check, CCC), combining diverse financial reasoning scenarios with verified annotations. Our models, DianJin-R1-7B and DianJin-R1-32B, are fine-tuned from Qwen2.5-7B-Instruct and Qwen2.5-32B-Instruct using a structured format that generates both reasoning steps and final answers. To further refine reasoning quality, we apply Group Relative Policy Optimization (GRPO), a reinforcement learning method that incorporates dual reward signals: one encouraging structured outputs and another rewarding answer correctness. We evaluate our models on five benchmarks: three financial datasets (CFLUE, FinQA, and CCC) and two general reasoning benchmarks (MATH-500 and GPQA-Diamond). \ignore{We adopt a structured training paradigm where models generate reasoning steps and final answers using supervised fine-tuning. To further improve reasoning quality, we use Group Relative Policy Optimization (GRPO)—a reinforcement learning algorithm that incorporates dual reward signals for output structure and answer accuracy. We evaluate our models, DianJin-R1-7B and DianJin-R1-32B, based on Qwen2.5-7B-Instruct and Qwen2.5-32B-Instruct, across three financial test sets including CFLUE, FinQA, and CCC, as well as two general benchmarks, MATH-500 and GPQA-Diamond. }Experimental results show that DianJin-R1 models consistently outperform their non-reasoning counterparts, especially on complex financial tasks. Moreover, on the real-world CCC dataset, our single-call reasoning models match or even surpass the performance of multi-agent systems that require significantly more computational cost. These findings demonstrate the effectiveness of DianJin-R1 in enhancing financial reasoning through structured supervision and reward-aligned learning, offering a scalable and practical solution for real-world applications.\ignore{ Evaluations across five challenging benchmarks (CFLUE, FinQA, CCC, MATH-500, and GPQA-Diamond) show that our reasoning-augmented models DianJin-R1-7B and DianJin-R1-32B based on Qwen2.5-7B-Struct and Qwen2.5-32B-Instruct significantly outperform non-reasoning baselines, with notable gains in complex financial tasks. We also demonstrate a practical multi-agent compliance system powered by DianJin-R1, showcasing its real-world applicability. Our work presents a scalable, domain-adapted strategy for advancing reliable and interpretable financial reasoning in LLMs.}
\end{abstract}

\section{Introduction}
Recent advances in large language models (LLMs) have led to growing interest in enhancing their reasoning abilities. Models such as OpenAI o1~\citep{openai-2024-reasoning}, DeepSeek R1~\citep{guo-etal-2025-deepseek-r1} and QwQ~\citep{qwen-2024-qwq} have shown that explicitly modeling reasoning processes can significantly boost performance on complex tasks~\citep{zhong-etal-2024-evaluation}. Despite these improvements, recent evaluations on financial benchmarks~\citep{xie-etal-2023-pixiu,xie-etal-2024-finben,zhu-etal-2024-benchmarking,chen-etal-2024-fintextqa,qian-etal-2025-fino1,liu-etal-2025-finr1} reveal that reasoning in this domain remains particularly challenging, given the need for domain-specific knowledge, accurate numerical reasoning, and strict compliance with regulatory requirements. Effectively addressing these challenges calls for specialized reasoning strategies capable of handling both structured financial information and open-ended problem solving. In response, we introduce DianJin-R1, LLMs that incorporate reasoning-augmented supervision and reinforcement learning to enhance performance on financial reasoning tasks.

We begin by constructing a high-quality reasoning dataset, DianJin-R1-Data, using three major sources: CFLUE~\citep{zhu-etal-2024-benchmarking}, FinQA~\citep{chen-etal-2021-finqa}, and our proprietary compliance dataset for the task of Chinese Compliance Check (CCC). CFLUE, which includes over 31,000 reasoning-annotated multiple-choice and open-ended questions from financial qualification mock exams, plays a central role in training due to its scale and diversity. FinQA provides numerical reasoning questions, while CCC focuses on complex compliance scenarios requiring multi-step logic. To ensure the quality of reasoning, we adopt a verification process using GPT-4o~\citep{openai-2024-gpt4o} to check for alignment between generated answers, reasoning steps, and reference explanations. This process results in a reliable set of reasoning-augmented and non-reasoning samples, supporting more robust model training.

For supervised fine-tuning (SFT), we train DianJin-R1-7B and DianJin-R1-32B, based on Qwen2.5-7B-Instruct and Qwen2.5-32B-Instruct~\citep{yang-etal-2024-qwen2.5}, to generate both the reasoning process and final answers using a structured output format with \texttt{<think>} and \texttt{<answer>} tags. To further improve reasoning quality, we apply Group Relative Policy Optimization (GRPO)~\citep{shao-etal-2024-deepseekmath}, a reinforcement learning algorithm that introduces two reward signals: a format reward to encourage structured outputs and an accuracy reward to promote answer correctness. These mechanisms guide the model to produce coherent, verifiable reasoning paths and reliable answers.

We evaluate our DianJin-R1 models, along with other general reasoning and non-reasoning models, across a diverse set of benchmarks, including CFLUE, FinQA, CCC, MATH-500~\citep{hendrycks-etal-2021-math}, and GPQA-Diamond~\citep{rein-etal-2024-gpqa}. The results demonstrate that reasoning-augmented models consistently outperform their non-reasoning counterparts, especially in the financial domain. Notably, training on CFLUE alone yields substantial gains across all tasks, and combining all datasets further enhances performance. Our analysis also highlights the benefit of reinforcement learning, particularly when the reward signals align with the task domain.

Finally, we demonstrate a practical application of our approach on the CCC dataset, where a multi-agent system based on LLMs is employed to perform condition-based compliance checks. By assigning specialized agents to each decision node in the workflow, the system effectively integrates intermediate reasoning steps to arrive at the final compliance judgment.

In summary, DianJin-R1 presents a scalable and effective strategy for enhancing financial reasoning in LLMs by combining high-quality supervision, structured reasoning generation, and reward-driven refinement through reinforcement learning.

\section{DianJin-R1-Data Construction}
\label{sec:dianjin-r1-data}

\subsection{Data Source}
\label{sec:data-source}

Our dataset originates from different sources: two open-source datasets and an in-house dataset. 

\paragraph{CFLUE~\citep{zhu-etal-2024-benchmarking}.} It is an open-source Chinese benchmark designed to assess the performance of LLMs on a variety of natural language processing (NLP) tasks within the financial domain. Its knowledge assessment component includes 38,638 multiple-choice financial exam questions, sourced from 15 types of financial qualification mock exams that cover various subjects and difficulty levels. To construct a high-quality subset for our study, we apply a three-step filtering process—focusing on question length, difficulty, and ambiguity. First, we apply a length filter to remove questions with fewer than 15 tokens, as these typically require minimal reasoning and offer limited value for assessing deeper understanding. \ignore{Second, in the difficulty filter, we discard overly simple questions—specifically those that are correctly answered by all smaller LLMs, including LLaMA-3.1-8B and Qwen2.5-7B-Instruct—to ensure the remaining questions demand more substantial reasoning.}Second, since simple QA pairs may not significantly enhance reasoning ability~\citep{ye-etal-2025-limo,muennighoff-etal-2025-s1}, we apply a difficulty filter to discard questions that are correctly answered by all smaller language models, including LLaMA-3.1-8B and Qwen2.5-7B-Instruct. This helps ensure that the remaining questions demand deeper reasoning. Finally, in the ambiguity filter, we use \texttt{GPT-4o}~\citep{openai-2023-gpt4} to eliminate questions that contain ambiguous wording, thereby ensuring each question is clear and well-defined. Through this filtering process, we curate a refined set of high-quality multiple-choice questions, each with an unambiguous correct answer, suitable for evaluating financial reasoning in LLMs. Notably, most of these question-answer pairs are accompanied by detailed explanations, which serve as valuable reasoning annotations.

\paragraph{FinQA~\citep{chen-etal-2021-finqa}.} FinQA is an open-source English benchmark containing 8,281 financial question-answer pairs that require numerical reasoning over financial reports. For our study, we apply the same length and difficulty filters as used in the CFLUE dataset to ensure quality and complexity. As a result, we curate a high-quality subset of QA pairs, well-suited for evaluating financial reasoning in English-language contexts.

\paragraph{CCC (Chinese Compliance Check).} It is an in-house dataset designed to detect compliance violations by service agents in real-world Chinese financial customer-service dialogues. Each instance in CCC includes the full conversation between a customer and a service agent, along with associated meta-information such as customer service actions (e.g., ticket escalation), and other relevant contextual data. The data is sourced from an online quality inspection system used in actual customer service operations, and each instance has been manually reviewed to ensure labeling accuracy. While compliance checking can be framed as a binary classification task (violation vs. non-violation), it is inherently more complex, as it requires evaluating whether the service agent’s behavior adheres to a set of domain-specific regulatory guidelines. Specifically, we sample from the manually validated data, ensuring a roughly balanced distribution between compliant and non-compliant cases.

\subsection{Reasoning Dataset Construction}
\label{sec:reasoning-path}

Considering the differences between the dataset CCC and the two others CFLUE and FinQA, we adopt different methods for reasoning construction for multiple-choice questions in CFLUE (Section~\ref{sec:reasoning-path-cflue}), QA pairs of FinQA (Section~\ref{sec:reasoning-path-finqa}) and dialogues of CCC (Section ~\ref{sec:reasoning-path-ccc}). Table~\ref{tab:stat} shows the statistics. 

\subsubsection{Reasoning Generation for CFLUE Questions}
\label{sec:reasoning-path-cflue}

We denote our selected multiple-choice questions from CFLUE as $D_{\text{CFLUE}_{\text{MCQ}}} = {(x_i, e_i, y_i)}$, where each $x_i$ includes both the question and its corresponding set of multiple-choice options, $e_i$ is the detailed explanation, and $y_i$ includes both a summarized reasoning process and the correct answer, typically indicated by a letter such as \textit{A}, \textit{B}, \textit{C}, or \textit{D}.\ignore{$y_i$ represents the correct answer—typically indicated as a letter (e.g., \textit{A}, \textit{B}, \textit{C}, or \textit{D}).}

\paragraph{Open-ended question generation.} We begin by using \texttt{GPT-4o} to convert each multiple-choice question in $D_{\text{CFLUE}_{\text{MCQ}}}$ into an open-ended format, where $x$ denotes the input question and $y$ represents the correct answer. Figure~\ref{fig:mcq2oe} in Appendix~\ref{apdx:converting} provides an illustration of this conversion process. We denote the resulting dataset of open-ended questions as $D_{\text{CFLUE}_{\text{OE}}} = \{(x_i, y_i)\}$. 

\paragraph{Reasoning generation.}
For each pair $(x_i, y_i)$ in $D_{\text{CFLUE}_{\text{MCQ}}}$, we leverage \texttt{DeepSeek-R1}~\citep{guo-etal-2025-deepseek-r1}, a model known for its strong reasoning capabilities, to generate a chain-of-thought (CoT) $r_i$ along with a predicted answer $a_i$ as follows:

\begin{equation} 
r_i, a_i = \text{DS}(x_i). 
\end{equation}

Next, we employ \texttt{GPT-4o} as a verifier to assess two key aspects of the generated output: (1) whether the predicted answer $a_i$ matches the gold answer $y_i$, and (2) whether the generated reasoning $r_i$ is consistent with the reference explanation $e_i$. If both conditions are satisfied, we retain the instance $(x_i, r_i, y_i)$ as a valid reasoning sample. If not, we retry the reasoning generation process, allowing up to $T$ attempts (with $T = 3$ in this paper). If all attempts fail to produce a correct answer and consistent reasoning, the instance $(x_i, y_i)$ is considered a hard case and preserved as a non-reasoning sample.  We denote the resulting reasoning-augmented dataset as $R_{\text{CFLUE}_{\text{MQC}}} = \{(x_i, r_i, y_i)\}$ and the hard, non-reasoning dataset as $G_{\text{CFLUE}_{\text{MQC}}} = \{(x_i, y_i)\}$.

We apply the same procedure to the open-ended question dataset $D_{\text{CFLUE}_{\text{OE}}}$, producing the corresponding reasoning-augmented dataset $R_{\text{CFLUE}_{\text{OE}}} = \{(x_i, r_i, y_i)\}$ and the hard, non-reasoning dataset as $G_{\text{CFLUE}_{\text{OE}}} = \{(x_i, y_i)\}$.

\subsubsection{Reasoning Generation for FinQA Questions}
\label{sec:reasoning-path-finqa}

Unlike the instances in CFLUE, the QA pairs in FinQA are already in an open-ended format. We denote the FinQA dataset similarly as $D_{\text{FinQA}} = \{(x_i, y_i)\}$.

We then apply the procedure of reasoning generation as open-ended questions in CFLUE to the QA pairs in $D_{\text{FinQA}}$. As a result, we obtain the reasoning-augmented dataset from FinQA as $R_{\text{FinQA}} = \{(x_i, r_i, y_i)\}$ and the hard, non-reasoning dataset as $G_{\text{FinQA}} = \{(x_i, y_i)\}$. 

\subsubsection{Reasoning Generation via Multi-Agent for CCC Dialogues}
\label{sec:reasoning-path-ccc}

\ignore{
\paragraph{Open-ended question generation.} For each instance in CCC, we wrap it as a open-ended question-answer pair. This is done by augmenting the dialogue with a question \textcolor{red}{\textit{**English**/**中文**}} and specifying the answer as \textcolor{red}{\textit{Yes/是的}} for having compliance violation and \textcolor{red}{\textit{No/不是}} for not having. We denote the resulting dataset of open-ended questions as $D_{\text{CCC}} = {(x_i, y_i)}$.

\paragraph{Reasoning generation via mulit-agent.}} We denote the CCC dataset as $D_{\text{CCC}} = \{(x_i, y_i)\}$, where $x_i$ is a dialogue and $y_i$ is the corresponding answer, which provides a summarized reasoning process and a final conclusion on whether a compliance violation has occurred. It is challenging for LLMs to directly generate reasoning from a dialogue $x_i$, since even humans typically rely on a set of guidelines to determine whether a service agent has violated compliance. To replicate this process, we have developed a workflow based on these guidelines, outlining the detailed steps for identifying compliance violations. The workflow begins with a start node and ends with two possible outcome nodes. One outcome indicates that the dialogue has no compliance violation, while the other indicates a compliance violation. All other nodes in the workflow are internal condition nodes, each evaluating whether a specific condition is met and triggering corresponding actions based on the result.

\begin{figure}[!ht]
\centering
\includegraphics[width=1.0\linewidth]{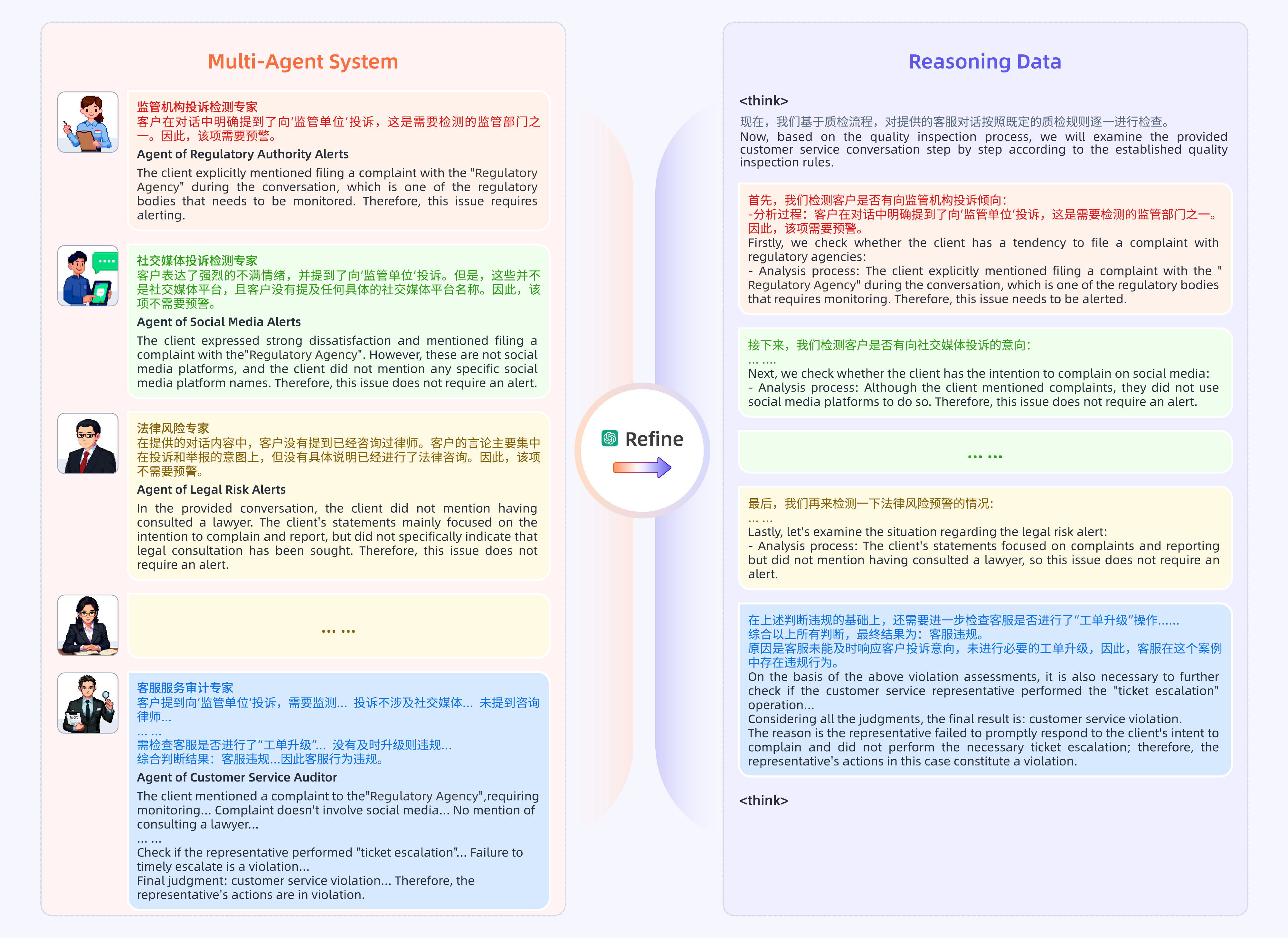}
\caption{An example of reasoning data synthesized by a multi-agent system.}
\label{fig:guidelines}
\end{figure}

Although using a multi-agent LLM-based system to identify compliance violations is a natural approach, it significantly increases inference costs due to multiple agent interactions. As an alternative, we use a multi-agent LLM system to generate reasoning. Specifically, we employ an LLM-based agent (i.e., \texttt{Qwen2.5-72B-Instruct} in this study) for each condition node. For each pair $\left(x_i, y_i\right)$ in $D_{\text{CCC}}$, we strictly follow the workflow and ask the corresponding agents to generate intermediate CoTs and the associated intermediate answers. The intermediate CoTs and intermediate answers are denoted as $\left(r_i^{0}, a_i^{0}, \cdots, r_i^{n_i}, a_i^{n_i}\right)$, where $n_i$ is the number of condition nodes involved before reaching the outcome node. The final answer, $a_i$, is determined by the outcome node. 

If the generated answer $a_i$ matches the gold answer $y_i$, we use \texttt{GPT-4o} to merge all the intermediate CoTs $\left(r_i^{0}, \cdots, r_i^{n_i}\right)$ into a final, unified CoT $r_i$ and retain the instance $\left(x_i, r_i, y_i\right)$. If the answer does not match, we retry the reasoning generation process up to $T$ attempts. Similarly, we denote the resulting reasoning-augmented dataset as $R_{\text{CC}}=\{\left(x_i, r_i, y_i\right)\}$ and the hard, non-reasoning dataset as $G_{\text{CC}}=\{\left(x_i, y_i\right)\}$. Figure~\ref{fig:guidelines} presents an example of how the final unified CoT is generated from the intermediate CoTs produced by the multi-agent system.

\begin{table}[!ht]
\centering
\small
\begin{tabular}{l|lllll}
\toprule
Dataset  &  Language & Size & \texttt{Q}$_{token}$ & \texttt{R}$_{token}$ & \texttt{A}$_{token}$ \\
\midrule
\rowcolor[gray]{0.8}
%\Block[c]{1-6}{\texttt{Used in SFT}}
\multicolumn{6}{c}{Used in SFT}\\
\midrule
CFLUE$_{MCQ}$ & Chinese & 26,672 & 134.85 & 807.42 & 95.71 \\ 
CFLUE$_{OE}$ & Chinese & 5,045 & 49.28 & 857.04 & 485.60 \\ 
FinQA & English & 4,851 & 1048.38 & 1576.91 & 148.42 \\
CCC & Chinese & 1,800 & 1695.78 & 884.29 & 69.64 \\
\midrule
\rowcolor[gray]{0.8}
\multicolumn{6}{c}{Used in RL}\\
\midrule
CFLUE$_{MCQ}$ & Chinese & 4096 & 132.40 & - & 2.15 \\ 
\bottomrule
\end{tabular}
\caption{Overview of datasets used in DianJin-R1-Data.}
\label{tab:stat}
\end{table}

\ignore{
\begin{table}[!ht]
\centering
\small
\begin{tabular}{l|l|llll|lll}
\toprule
\multirow{2}{*}{\bf Dataset}  & \multirow{2}{*}{\bf Language} & \multicolumn{4}{c|}{\bf Used in SFT}  & \multicolumn{3}{c}{\bf Used in RL}\\
\cmidrule{3-6}\cmidrule{7-9}
& & \bf Size & \bf \texttt{Q}$_{token}$ & \bf \texttt{R}$_{token}$ & \bf \texttt{A}$_{token}$ & \bf Size & \bf \texttt{Q}$_{token}$ & \bf \texttt{A}$_{token}$ \\
\midrule
CFLUE$_{MCQ}$ & Chinese & 26,672 & 134.85 & 807.42 & 95.71 & 4096 & 132.40 & 2.15 \\ 
CFLUE$_{OE}$ & Chinese & 5,045 & 49.28 & 857.04 & 485.60 & - & - & - \\ 
FinQA & English & 4,851 & 1048.38 & 1576.91 & 148.42 & - & - & -  \\
CCC & Chinese & 1,800 & 1695.78 & 884.29 & 69.64 & - & - & - \\
\midrule
All & - & 38368 & 312.32 & 914.84 & 152.41 & - & - & - \\
\bottomrule
\end{tabular}
\caption{: Overview of datasets used in DianJin-R1-Data.}
\label{tab:stat}
\end{table}
}

\subsection{Model Training}
\begin{figure}[!ht]
    \centering
    \includegraphics[width=1.0 \columnwidth]{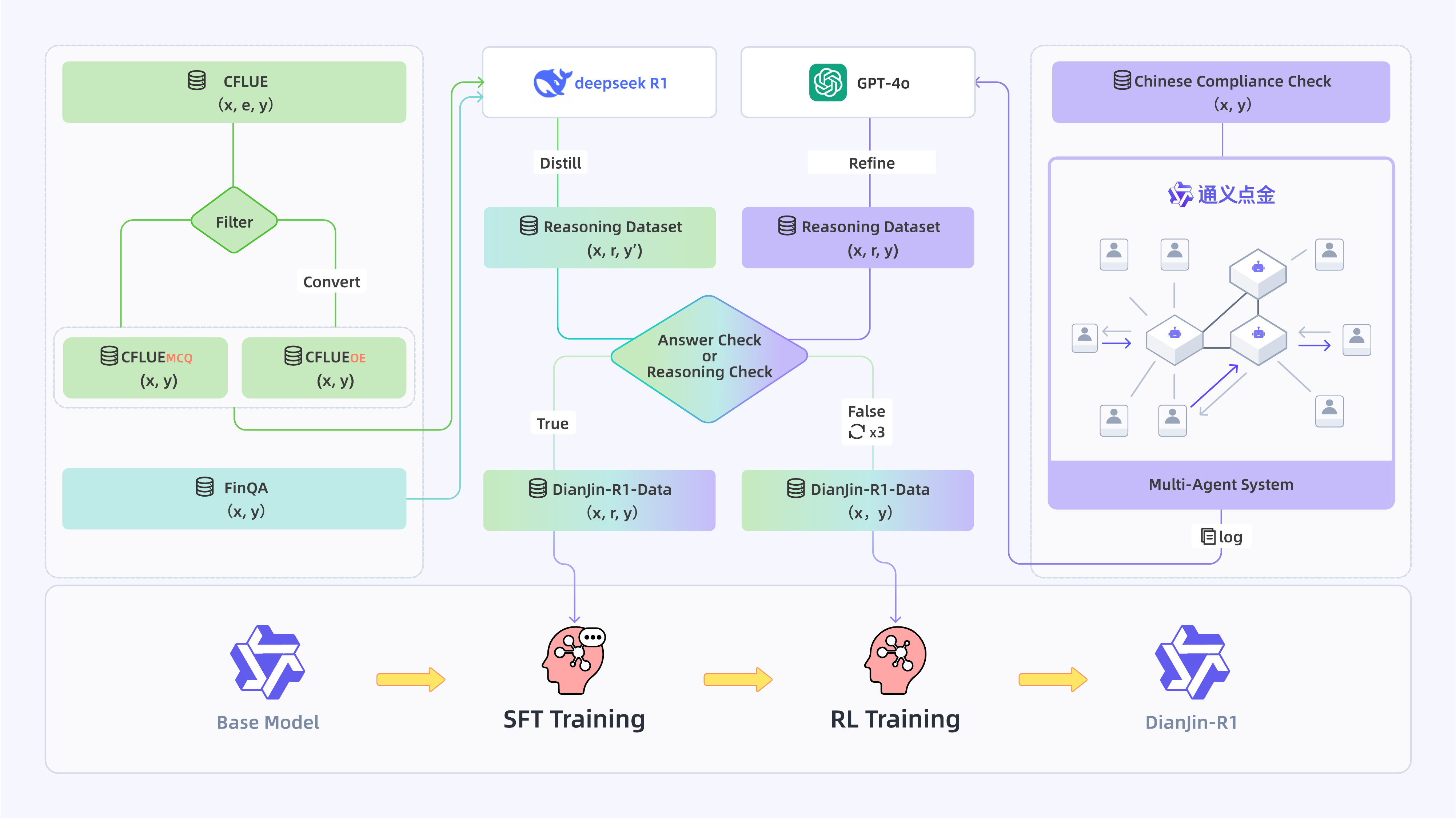}
    \caption{Illustration of two-step training for DianJin-R1.}
    \label{fig:training}
\end{figure}

In this section, we present the details for training LLMs to perform financial reasoning. As shown in Figure~\ref{fig:training}, the training consists of two stages: learning reasoning with supervised fine-tuning (SFT) and enhancing reasoning with reinforcement learning (RL).

\subsubsection{Learning Reasoning with SFT} 

The reasoning datasets—$R_{\text{CFLUE}_{\text{MCQ}}}$, $R_{\text{CFLUE}_{\text{OE}}}$, $R_{\text{FinQA}}$, and $R_{\text{CCC}}$—are utilized to fine-tune LLMs to generate a chain-of-thought (CoT) followed by a final answer. Each training instance $(x, r, y)$ consists of a question $x$, a reasoning path $r$ (formatted as \texttt{<think>}$\cdots$\texttt{</think>}), and an answer $y$ (formatted as \texttt{<answer>}$\cdots$\texttt{</answer>}). During fine-tuning, the question $x$ serves as the input to the model, while the reasoning $r$ and final answer $y$ are treated as the target output, enabling the model to learn to produce coherent reasoning steps along with the correct solution.

\subsubsection{Enhancing Reasoning with RL}
The hard and non-reasoning dataset $G_{\text{CFLUE}_{\text{MCQ}}}$\ignore{s, including $G_{\text{CFLUE}{\text{MCQ}}}$, $G_{\text{CFLUE}_{\text{OE}}}$, $G_{\text{FinQA}}$, and $G_{\text{CCC}}$ are} is used in the reinforcement learning (RL), which aims to further enhance the reasoning skills.\footnote{In the future, we will include $G_{\text{CFLUE}_{\text{OE}}}$, $G_{\text{FinQA}}$, and $G_{\text{CCC}}$ in RL.} Specifically, we adopt the Group Relative Policy Optimization (GRPO) algorithm~\citep{shao-etal-2024-deepseekmath} for RL, incorporating two reward mechanisms: a format reward to ensure the generated output adheres to the desired structure, and an accuracy reward to encourage correct answers.

\begin{itemize}
    \item Format reward: We incorporate a format reward to promote well-structured outputs. For a given response $y$, a reward score of 1 is granted if the output contains exactly one reasoning segment enclosed within \texttt{<think>}$\cdots$\texttt{</think>} tags and one final answer enclosed within \texttt{<answer>}$\cdots$\texttt{</answer>} tags, with no additional content outside these boundaries. If this strict formatting criterion is not met, the output receives a reward score of 0.
    \item Accuracy reward: We use an accuracy reward to promote answer correctness. Specifically, for a given output $y$, if the content enclosed within the \texttt{<answer>}$\cdots$\texttt{</answer>} tags exactly matches the reference answer,\footnote{In the current version, the accuracy reward is applied exclusively to multiple-choice questions from $R_{\text{CFLUE}_{\text{MCQ}}}$.} the model receives a reward score of 1; otherwise, it receives a score of 0. This mechanism encourages the generation of accurate and reliable final answers.
\end{itemize}

\section{Experimentation}

\subsection{Experimental Setups}
\paragraph{Training Data.} Table~\ref{tab:stat} summarizes the statistics for DianJin-R1-Data. Since \text{CFLUE}$_{\text{MCQ}}$ constitutes a large portion of the SFT data, we shuffle it together with the other datasets to prevent overfitting to any single source.

\paragraph{Model Training.} Using the constructed DianJin-R1-Data and our proposed training method, we train DianJin-R1-7B and DianJin-R1-32B based on Qwen2.5-7B-Instruct and Qwen2.5-32B-Instruct, respectively. Our model training consists of two stages,conducted on NVIDIA A100 GPUs. In the SFT stage, the 7B model is trained on a single node with 8 GPUs, and the 32B model uses 32 GPUs across 4 nodes. We leverage DeepSpeed's Zero-3 for optimization, using a learning rate of $1.0\times10^{-5}$, a sequence length of 16K, and bf16 precision over 3 epochs. We apply gradient accumulation with 16 steps to simulate a larger batch size. In the RL stage, we perform 8 rollouts per sample, with a train batch size of 1024 and a rollout batch size of 256. We use a learning rate of $1.0\times10^{-6}$ and a sampling temperature of 0.6 over 5 epochs to balance learning dynamics.

\paragraph{Evaluation Datasets.} We evaluate our models on three financial benchmarks, including the test sets of CFLUE, FinQA, and our proprietary in-house dataset, CCC. To further assess general reasoning capabilities, we also include two widely-used benchmarks: MATH-500~\citep{hendrycks-etal-2021-math} and GPQA-Diamond~\citep{rein-etal-2024-gpqa}. For each dataset, we report the accuracy—defined as the proportion of correctly answered questions—and compute the average accuracy across all test sets. Among them, CFLUE and CCC are Chinese-language datasets, while the others are in English. The detailed statistics of these test sets are summarized in Table~\ref{tab:test_stat}. For FinQA and CCC, we use GPT-4o to evaluate the correctness of each answer, following the prompts shown in Figure~\ref{fig:prompt_judge_fin_qa} and Figure~\ref{fig:prompt_judge_ccc} in Appendix~\ref{apdx:prompts}. For the other test sets, we extract the predicted answers using rule-based methods and compare them directly with the gold answers.

\begin{table}[!ht]
\centering
\small
\begin{tabular}{l|ll}
\toprule
\bf Dataset & \bf Language & \bf Size \\
\midrule
\rowcolor[gray]{0.8}
\multicolumn{3}{c}{Financial domain}\\
\midrule
CFLUE  & Chinese & 3,864 \\
FinQA & English & 1,147 \\
CCC & Chinese & 200 \\
\midrule
\rowcolor[gray]{0.8}
\multicolumn{3}{c}{General domain}\\
\midrule
MATH-500 & English & 500 \\
GPQA-Diamond & English & 198 \\
\bottomrule
\end{tabular}
\caption{Statistics of the test sets.}
\label{tab:test_stat}
\end{table}

\paragraph{Baselines} We compare our models against two categories of LLMs. The first includes general LLMs without explicit reasoning capabilities: GPT-4o~\citep{openai-2024-gpt4o}, DeepSeek-V3~\citep{liu-etal-2024-deepseek-v3}, Qwen2.5-7B-Instruct~\citep{yang-etal-2024-qwen2.5}, Qwen2.5-32B-Instruct, and Qwen2.5-72B-Instruct. The second category consists of general LLMs equipped with reasoning abilities, including DeepSeek-R1~\citep{guo-etal-2025-deepseek-r1}, DeepSeek-R1-Distill-Qwen-7B, DeepSeek-R1-Distill-Qwen-14B, DeepSeek-R1-Distill-Qwen-32B, and QwQ-32B~\citep{qwen-2024-qwq}.

\subsection{Experimental Results}

Table~\ref{tab:result} compares the performance of LLMs with and without explicit reasoning. Overall, the results demonstrate that models incorporating reasoning generally outperform their non-reasoning counterparts, with the exception of DeepSeek-R1-Distill-Qwen-7B. The results reveal the following key points:
\begin{itemize}
\item On the three financial test sets (i.e., CFLUE, FinQA, and CCC), our DianJin-R1 models significantly outperform the base models (Qwen2.5-7B-Instruct and Qwen2.5-32B-Instruct), especially on CFLUE and CCC. For instance, DianJin-R1-32B improves accuracy from 77.95 to 86.74 on CFLUE, from 79.51 to 80.82 on FinQA, and from 56.50 to 96.00 on CCC. Encouragingly, DianJin-R1-32B achieves the highest accuracy on these financial tasks, surpassing even the performance of DeepSeek-R1, a model known for its strong reasoning capabilities. 
\item On the two general-domain test sets (i.e., MATH-500 and GPQA-Diamond), we also observe performance improvements in the DianJin-R1 models compared to their respective base models. For example, DianJin-R1-32B improves accuracy from 81.00 to 88.20 on MATH-500, and from 44.95 to 58.59 on GPQA-Diamond. This suggests that training on financial reasoning data can enhance general reasoning capabilities to some extent. However, since our SFT and RL training pipelines do not incorporate any general-domain reasoning datasets, the performance of DianJin-R1 models on these benchmarks remains lower than that of models with larger parameter sizes or those fine-tuned on general reasoning data.
\item Although general-purpose reasoning models such as DeepSeek-R1 and QwQ-32B significantly improve performance on general reasoning benchmarks like MATH-500 and GPQA-Diamond, they do not always yield better results on financial benchmarks. For instance, while DeepSeek R1 performs better than DeepSeek-V3 on CFLUE and CCC, it actually leads to a drop in performance on FinQA. Similarly, DeepSeek-R1-Distill-Qwen-7B performs worse than Qwen-2.5-7B-Instruct across all financial test sets. These findings are similar to those previously reported in Fino1~\citep{qian-etal-2025-fino1}.
\end{itemize}

\begin{table}[!ht]
\centering
\small
\begin{tabular}{l|lllll|l}
\toprule
\multirow{2}{*}{Model} & \multicolumn{3}{c}{Financial} & \multicolumn{2}{c|}{General} & \multirow{2}{*}{Avg.}\\
\cmidrule(lr){2-4}\cmidrule(lr){5-6}
& CFLUE & FinQA & CCC & MATH\ignore{-500} & GPQA\ignore{-Diamond} \\
\midrule
\rowcolor[gray]{0.9}
\multicolumn{7}{c}{General models without explicit reasoning}\\
\hline
GPT-4o & 71.68 & 79.16 & 50.00 & 77.93 & 39.56 & 63.67  \\
DeepSeek-V3 & 75.14 & \bf{81.34} & 57.50 & 87.20 & 45.45 & 68.33 \\
Qwen2.5-7B-Instruct & 69.37 & 66.70 & 55.00 & 71.40 & 33.84 & 59.26 \\
Qwen2.5-32B-Instruct & 77.95 & 79.51 & 56.50 & 81.00 & 44.95 & 67.98 \\
Qwen2.5-72B-Instruct & 79.46 & 77.94 & 55.50 & 82.20 & 39.90 & 67.00 \\
\midrule
\rowcolor[gray]{0.9}
\multicolumn{7}{c}{General models with reasoning}\\
\hline
DeepSeek-R1 & \underline{86.64} & 79.81 & 67.50 & \underline{94.80} & \bf{66.16} & \underline{78.98} \\
DeepSeek-R1-Distll-Qwen-7B & 48.39 & 66.09 & 41.50 & 90.20 & 45.96 & 58.43 \\
DeepSeek-R1-Distill-Qwen-14B & 70.83 & 76.63 & 50.00 & 93.20 & 54.55 & 69.04 \\
DeepSeek-R1-Distill-Qwen-32B & 78.52 & 77.00 & 52.00 & \bf{95.00} & \underline{63.64} & 73.23 \\
QwQ-32B & 83.49 & 78.38 & 52.00 & \bf{95.00} & \underline{63.64} & 74.50 \\
\midrule
\rowcolor[gray]{0.9}
\multicolumn{7}{c}{DianJin-R1 with reasoning}\\
\hline
DianJin-R1-7B & 80.32 & 77.72 & \underline{94.50} & 76.60 & 37.54 & 73.34 \\
DianJin-R1-32B & \bf{86.74} & \underline{80.82} & \bf{96.00} & 88.20 & 58.59 & \bf{82.07}  \\
\bottomrule
\end{tabular}
\caption{Performance comparison in accuracy across different test sets. Scores in \textbf{bold} and \underline{underlined} indicate the best and second-best results, respectively.}
\label{tab:result}
\end{table}

\subsection{Discussion}
In this section, we use \texttt{Qwen2.5-7B-Instruct} as the backbone model to investigate the impact of RL and different datasets used during SFT.

% \paragraph{Effect of different methods in RL.} 

% \begin{table}[!ht]
% \centering
% \small
% \begin{tabular}{l|llllll|l}
% \toprule
% \multirow{2}{*}{RL} & \multicolumn{3}{c}{Financial} & \multicolumn{3}{c|}{General} & \multirow{2}{*}{Avg.}\\
% \cmidrule(lr){2-4}\cmidrule(lr){5-7}
% & CFLUE & FinQA & CCC & MATH-500 & GPQA\ignore{-Diamond} & AIME-2024 & \\
% \midrule
% None & 75.48 & 80.28 & 93.00 & 74.87 & 34.85 & 13.33 \\
% PPO & 76.34 & 75.24 & 95.00 & 76.93 & 38.89 & 20.00 \\
% GPRO & \bf 76.63 & 76.20 & 95.00  & 75.93 & 36.70 & 13.33 \\
% \bottomrule
% \end{tabular}
% \caption{Performance comparison across different RL methods. Here, \textit{None} indicates that no RL is applied.}
% \label{tab:rl-effect}
% \end{table}

\paragraph{Effect of RL.}
As shown in Table~\ref{tab:rl-effect}, SFT significantly enhances performance across all datasets. This indicates that SFT effectively equips the language models with reasoning capabilities. Furthermore, RL brings additional improvements on all datasets except FinQA. We suspect this exception may be due to the fact that all instances used for RL are in Chinese and sourced from CFLUE, whereas FinQA is in English. We plan to explore this further by incorporating English examples into the RL stage in future work.

\begin{table}[!ht]
\centering
\small
\begin{tabular}{l|lllll|l}
\toprule
\multirow{2}{*}{Learning} & \multicolumn{3}{c}{Financial} & \multicolumn{2}{c|}{General} & \multirow{2}{*}{Avg.}\\
\cmidrule(lr){2-4}\cmidrule(lr){5-6}
& CFLUE & FinQA & CCC & MATH\ignore{-500} & GPQA\ignore{-Diamond} & \\
\midrule
- & 69.37 & 66.70 & 55.00 & 71.40 & 33.84 & 59.26 \\
\midrule
SFT & 75.48 & \bf 80.28 & 93.00 & 74.87 & 34.85 & 71.70 \\
SFT + RL & \bf 80.32 & 77.72 & \bf 94.50 & \bf 76.60 & \bf 37.54 & \bf 73.34 \\
\bottomrule
\end{tabular}
\caption{Performance comparison in accuracy.}
\label{tab:rl-effect}
\end{table}

\paragraph{Effect of different datasets used in SFT.}
We use three data sources for SFT: CFLUE, FinQA, and CCC. To understand the individual contribution of each dataset, we evaluate model performance using different combinations of these sources. Table~\ref{tab:dataset-effect} presents the results. Among the datasets, CFLUE (which includes both CFLUE${_\text{MCQ}}$ and CFLUE${_\text{OE}}$) has the greatest impact. This is largely due to its large scale, as it contains more than 31,000 reasoning instances. When used alone, CFLUE significantly improves performance across all test sets, increasing overall accuracy from 59.26 to 65.67. Adding FinQA or CCC on top of CFLUE mainly enhances performance on their respective test sets, with limited influence on the others. Finally, using all three datasets together during SFT results in the best overall performance.

\begin{table}[!ht]
\centering
\small
\begin{tabular}{l|lllll|l}
\toprule
\multirow{2}{*}{Dataset} & \multicolumn{3}{c}{Financial} & \multicolumn{2}{c|}{General} & \multirow{2}{*}{Avg.}\\
\cmidrule(lr){2-4}\cmidrule(lr){5-6}
& CFLUE & FinQA & CCC & MATH\ignore{-500} & GPQA\ignore{-Diamond} & \\
\midrule
- & 69.37 & 66.70 & 55.00 & 71.40 & 33.84 & 59.26 \\
\midrule
CFLUE & 75.09 & 71.65 & 70.50 & 75.07 & \bf 36.03 & 65.67  \\
CFLUE + FinQA & 75.20 & \bf 80.85 & 70.00 & 75.33 & \bf 36.03 & 67.48  \\
CFLUE + CCC & 75.16 & 71.78 & \bf 95.00 & \bf 76.27 & 35.35 & 70.71  \\
CFLUE + CCC + FinQA & \bf 75.48 & 80.28 & 93.00 & 74.87 & 34.85 & \bf 71.70 \\
\bottomrule
\end{tabular}
\caption{Performance comparison in accuracy when different datasets are used in SFT. Here, RL is not applied.}
\label{tab:dataset-effect}
\end{table}

\paragraph{Effect of different systems on CCC.} As discussed earlier, it is natural to build a multi-agent LLM-based system to detect compliance violations in the CCC dataset. To achieve this, we follow the same reasoning generation workflow described in Section~\ref{sec:reasoning-path-ccc}. Specifically, we assign an LLM agent (i.e., \texttt{Qwen2.5-72B-Instruct}), to each condition node in the workflow. Each agent is responsible for predicting the intermediate outcome of its corresponding condition. The final decision is then derived from the result at the outcome node based on the collective outputs of all agents. Table~\ref{tab:cc-result} presents the performance comparison on the CCC test set. Among the non-reasoning systems, introducing the multi-agent approach alone significantly improves accuracy from 55.50 to 95.00, highlighting the effectiveness of structured, condition-based reasoning for this task. However, this comes at a high cost, as it requires an average of 8.15 API calls per instance. Notably, our reasoning-based models, DianJin-R1-7B and DianJin-R1-32B, achieve comparable or even superior performance with just a single API call despite their smaller sizes. This demonstrates the strength of our approach in learning to reason effectively and organizing reasoning paths to handle complex compliance evaluations.

\begin{table}[!ht]
\centering
\small
\begin{tabular}{l|ll}
\toprule
\bf System & \bf Accuracy & \bf \#Calls\\
\midrule
Qwen2.5-72B-Instruct & 55.50 & 1\\
Qwen2.5-72B-Instruct (Multi-Agent) & 95.00 & 8.15 \\
\midrule
DianJin-R1-7B & 94.50 & 1\\
DianJin-R1-32B & \bf 96.00 & 1 \\
\bottomrule
\end{tabular}
\caption{Performance comparison in accuracy on the test set of CCC. \#Calls denotes the average number of API calls required per test instance.}
\label{tab:cc-result}
\end{table}

% \section{Related Work}

\section{Conclusion and Future Work}
We have presented DianJin-R1, a reasoning-augmented framework for large language models in the financial domain. This framework combines structured supervision with a reinforcement learning algorithm (GRPO) to enhance model performance on complex financial and compliance-related tasks. Through extensive experiments on diverse benchmarks and a real-world compliance system, we demonstrate that reasoning-aware training significantly improves both accuracy and interpretability. 

In future work, we plan to explore alternative reinforcement learning strategies, including fine-grained reward shaping and hierarchical policy learning, to further refine reasoning quality. Moreover, we aim to incorporate tool-augmented reasoning, enabling models to dynamically invoke external tools—such as calculators, retrieval systems, or rule engines—during inference, with the goal of improving precision and robustness in high-stakes financial applications.

\section*{Ethics Statement}
We will open-source DianJin-R1-Data, including enhanced versions of CFLUE and Fin-QA datasets, and two model sizes, DianJin-R1-7B and DianJin-R1-32B. However, due to sensitivity concerns, CCC scenario data will not be made publicly available. The CCC tasks are designed solely to demonstrate the effectiveness of using a multi-agent system to synthesize reasoning data, and their exclusion does not affect the reproducibility of our methodology.

\section*{Acknowledgements}
We extend our sincere thanks to other members of the Qwen DianJin Team for their significant contributions, whose hard work and dedication were crucial to the success of this project. This work was supported by the Alibaba Innovative Research Program.

\bibliography{colm2024_conference}
\bibliographystyle{colm2024_conference}

\appendix

\section{Example of converting a multiple-choice question into an open-ended problem}
\label{apdx:converting}

\begin{figure}[t]
    \centering
    \includegraphics[width=1.0\columnwidth]{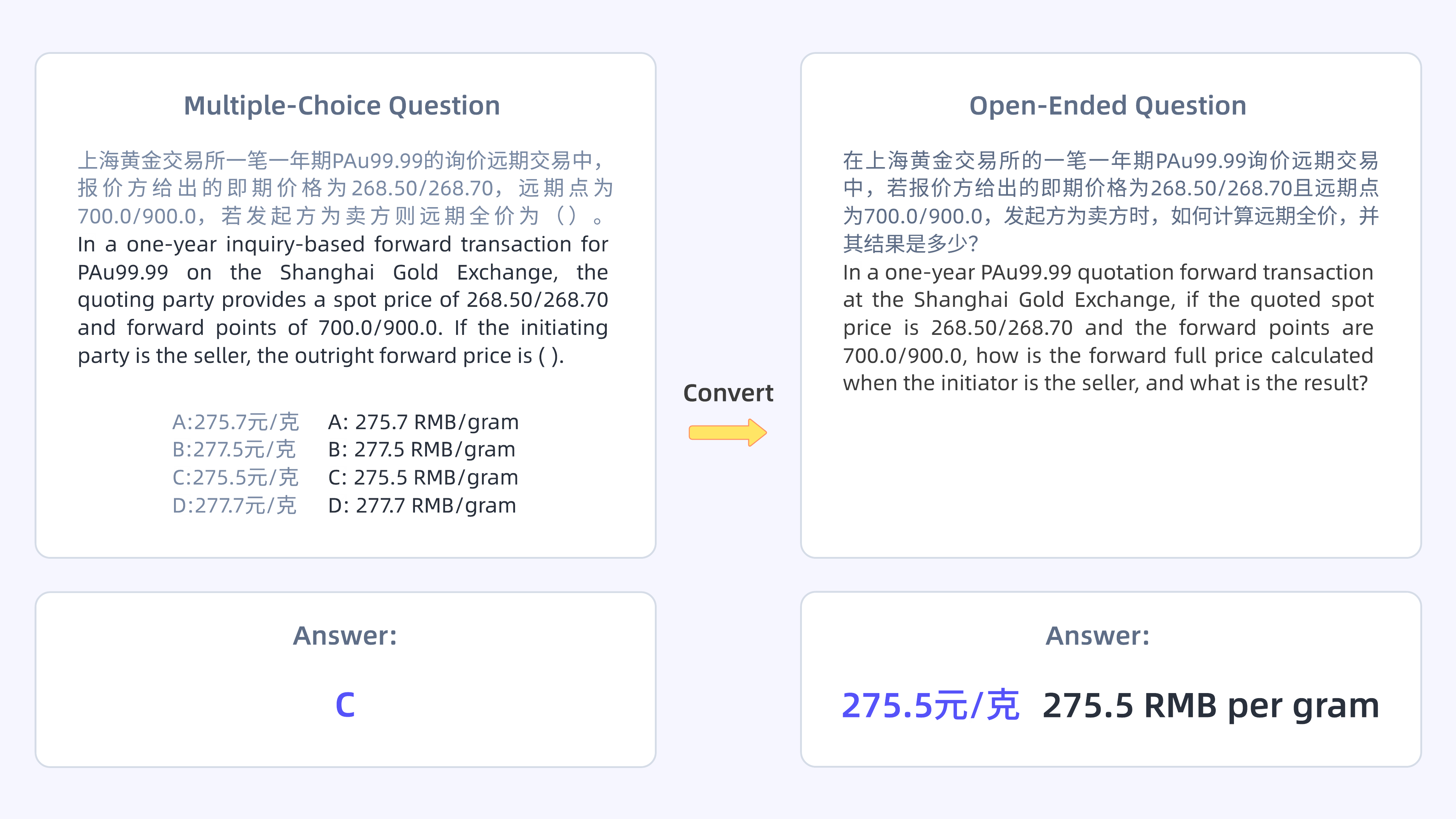}
    \caption{An example of converting a multiple-choice question from CFLUE into an open-ended format.}
    \label{fig:mcq2oe}
\end{figure}

Figure~\ref{fig:mcq2oe} illustrates an example of converting a multiple-choice question into an open-ended question. 

%\section{Example of open-ended question from CCC}

%\section{Guidelines and workflow in determining compliance violation}
%\label{apdx:guidelines-workflow}

\section{Prompts used in this paper}
\label{apdx:prompts}

Figure~\ref{fig:prompt_cflue2oe} illustrates the prompt used to convert a multiple-choice question from CFLUE into an open-ended format. Since the CFLUE test set includes both single-answer and multiple-answer multiple-choice questions, we design separate prompts for each type. The prompts for single-answer and multiple-answer questions are shown in Figure~\ref{fig:promt_cflue_single} and Figure~\ref{fig:prompt_cflue_mcq}, respectively.

\begin{figure*}[!t]
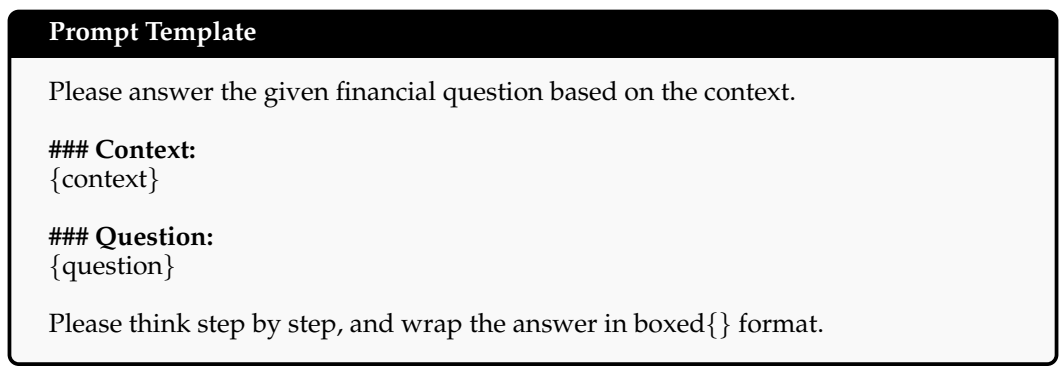

\begin{tcolorbox}[title=Prompt Template, colback=gray!5, colframe=black, fonttitle=\bfseries]
Please answer the given financial question based on the context.\\

\textbf{\#\#\# Context:} \\
\{context\}
\\

\textbf{\#\#\# Question:} \\
\{question\}
\\

Please think step by step, and wrap the answer in boxed\{\} format. 

\end{tcolorbox}
\caption{Prompt used to generate answers for questions in FinQA.}
\label{fig:prompt_fin_qa}
\end{figure*}

Figure~\ref{fig:prompt_fin_qa} presents the prompt used to generate answers for FinQA questions.

\begin{figure*}[!t]
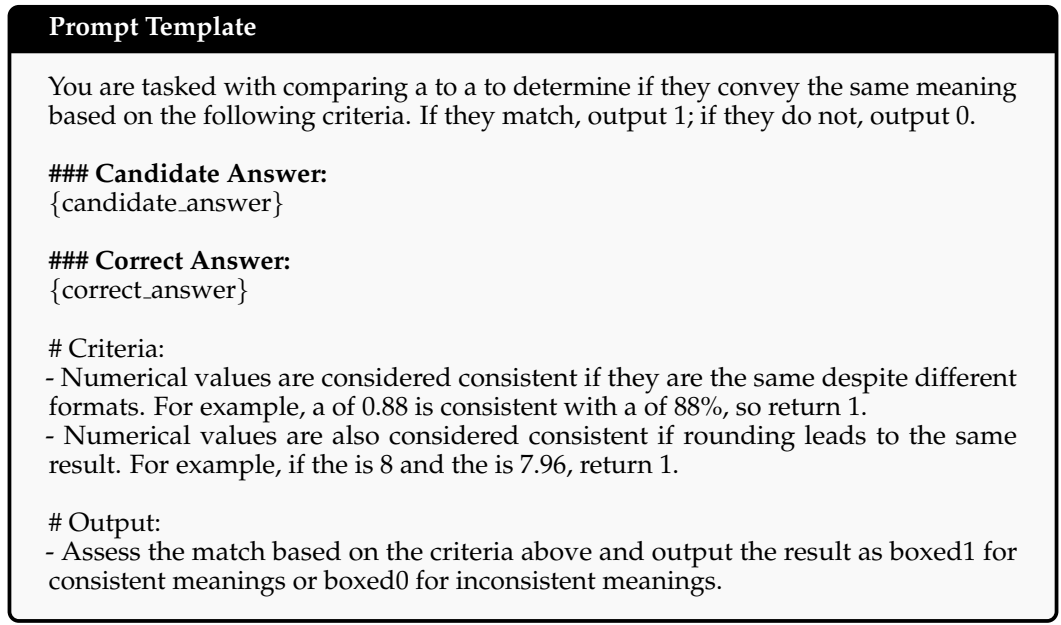

\begin{tcolorbox}[title=Prompt Template, colback=gray!5, colframe=black, fonttitle=\bfseries]
You are tasked with comparing a to a to determine if they convey the same meaning based on the following criteria. If they match, output 1; if they do not, output 0. \\

\textbf{\#\#\# Candidate Answer:} \\
\{candidate\_answer\} \\

\textbf{\#\#\# Correct Answer:}  \\
\{correct\_answer\} \\

\# Criteria: \\
- Numerical values are considered consistent if they are the same despite different formats. For example, a of 0.88 is consistent with a of 88\%, so return 1. \\
- Numerical values are also considered consistent if rounding leads to the same result. For example, if the is 8 and the is 7.96, return 1. \\

\# Output: \\
- Assess the match based on the criteria above and output the result as boxed{{1}} for consistent meanings or boxed{{0}} for inconsistent meanings.

\end{tcolorbox}
\caption{Prompt used to assess the correctness of model-generated answers in FinQA.}
\label{fig:prompt_judge_fin_qa}
\end{figure*}

\begin{figure*}[!t]
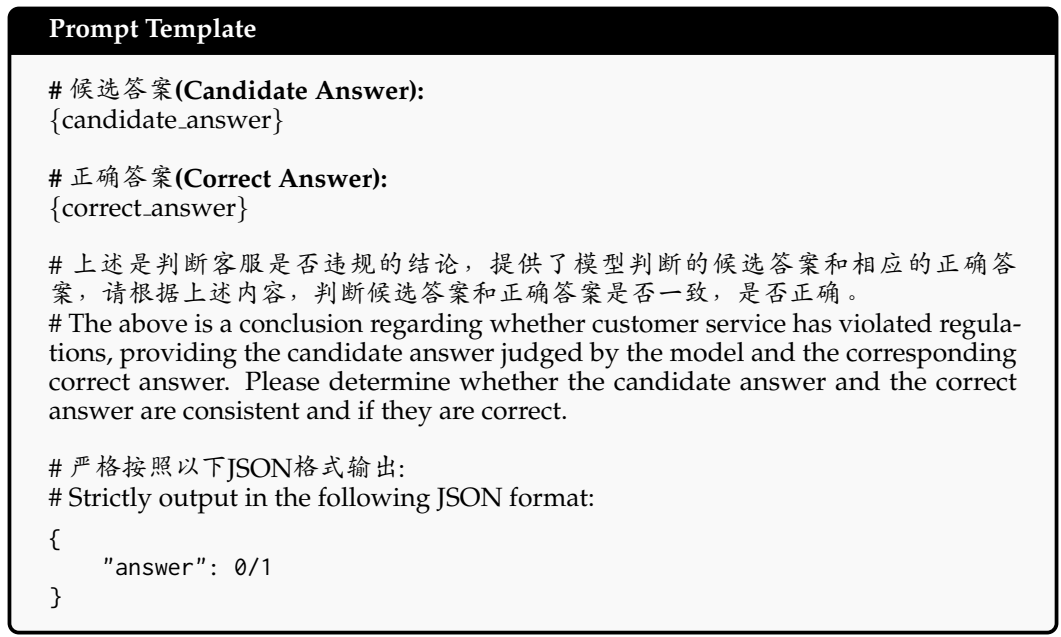

\begin{tcolorbox}[title=Prompt Template, colback=gray!5, colframe=black, fonttitle=\bfseries]
\textbf{\# 候选答案 (Candidate Answer):} \\
\{candidate\_answer\} \\

\textbf{\# 正确答案 (Correct Answer):}  \\
\{correct\_answer\} \\

\# 上述是判断客服是否违规的结论，提供了模型判断的候选答案和相应的正确答案，请根据上述内容，判断候选答案和正确答案是否一致，是否正确。 \\
\# The above is a conclusion regarding whether customer service has violated regulations, providing the candidate answer judged by the model and the corresponding correct answer. Please determine whether the candidate answer and the correct answer are consistent and if they are correct. \\

\# 严格按照以下JSON格式输出: \\
\# Strictly output in the following JSON format: 
\begin{verbatim}
{
    "answer": 0/1
}
\end{verbatim}

\end{tcolorbox}
\caption{Prompt used to assess the correctness of model-generated answers in CCC.}
\label{fig:prompt_judge_ccc}
\end{figure*}

Figure~\ref{fig:prompt_judge_fin_qa} and Figure~\ref{fig:prompt_judge_ccc} show the prompts used to evaluate the correctness of answers for questions in FinQA and CCC, respectively.

% Prompt used to generate reasoning for open-ended questions from CFLUE.
\begin{figure*}[!t]
\begin{tcolorbox}[title=Prompt Template, colback=gray!5, colframe=black, fonttitle=\bfseries]
我将向你提供一个多选题，你的任务是将其改成为一个开放式问题，并提供标准答案。要求如下：\\
I will provide you with a multiple-choice question, and your task is to rewrite it into an open-ended question and provide a standard answer. The requirements are as follows: \\
\\

- 问题必须具体，针对原多选题中测试的要点。确保问题是开放式的，即不提供选项，但必须有明确的标准答案。\\
- The question must be specific and address the key points tested in the original multiple-choice question. Ensure the question is open-ended, meaning no options are provided, but there must be a clear standard answer. \\
- 根据原问题的正确答案，提供一个简洁的标准答案。答案应当允许精确匹配，以确定模型的回答是否正确。\\
- Provide a concise standard answer based on the correct answer from the original question. The answer should allow for exact matching to determine if the model's response is correct. \\

\textbf{\#\#\# Multiple-choice Question} \\
\{question\} \\
\{options\} \\
\textbf{\#\#\# Correct Answer} \\
\{correct\_answer\} \\
\\
\#\# 严格按照以下JSON格式输出：\\
\#\# Strictly output in the following JSON format: \\

\begin{verbatim}
{
    "question": "" # Open-ended Question
    "answer": "" # Standard Answer
}
\end{verbatim}

\end{tcolorbox}
\caption{Prompt used to convert a multiple-choice question from CFLUE into an open-ended question.}
\label{fig:prompt_cflue2oe}
\end{figure*}

% CFLUE single prompt
\begin{figure*}[!t]
\begin{tcolorbox}[title=Prompt Template, colback=gray!5, colframe=black, fonttitle=\bfseries]
假设你是一位金融行业专家，请回答下列问题。\\
You are an expert in the finance industry, please answer the following questions. \\
注意：题目是单选题，只需要返回一个最合适的选项，若有多个合适的答案，只返回最准确的即可。\\
Note: The questions are multiple-choice, and you only need to return the most suitable option. If there are multiple suitable answers, return the most accurate one. \\

\textbf{\#\#\# Question:} \\
\{question\}
\\

\textbf{\#\#\# Choices:} \\
\{choices\}
\\

请一步步思考，并把答案选项放到 boxed\{\} 中。 \\
Please think step by step, and wrap the answer in boxed{} format. 

\end{tcolorbox}
\caption{Prompt used to generate answers for single-answer multiple-choice questions in CFLUE.}
\label{fig:promt_cflue_single}
\end{figure*}

% CFLUE MCQ prompt
\begin{figure*}[!t]
\begin{tcolorbox}[title=Prompt Template, colback=gray!5, colframe=black, fonttitle=\bfseries]
假设你是一位金融行业专家，请回答下列问题。\\
You are an expert in the finance industry, please answer the following questions. \\
注意：题目是多选题，可能存在多个正确的答案。\\
Note: The questions are multiple-choice and there may be multiple correct answers. \\

\textbf{\#\#\# Question:} \\
\{question\}
\\

\textbf{\#\#\# Choices:} \\
\{choices\}
\\

请一步步思考，并把答案选项放到 boxed\{\} 中。 \\
Please think step by step, and wrap the answer in boxed{} format. 

\end{tcolorbox}
\caption{Prompt used to generate answers for multiple-answer multiple-choice questions in CFLUE.}
\label{fig:prompt_cflue_mcq}
\end{figure*}

% FinQA

% Judge prompt for Fin-QA

% Judge prompt for CCC

\end{CJK*}
\end{document}